%% file: main.tex
\documentclass[10pt,twocolumn,letterpaper]{article}
\usepackage[pagenumbers]{cvpr}
\input{preamble}

\definecolor{cvprblue}{rgb}{0.21,0.49,0.74}
\usepackage[pagebackref,breaklinks,colorlinks,allcolors=cvprblue]{hyperref}
\usepackage[capitalize]{cleveref}
\title{GRUN: Generalized Retinex Unfolding Network for Image Restoration}
\author{Rihan Zhang, Fengyang Xiao, Yuhan Chen, Chunming He$^{*}$, and Sina Farsiu\\
Duke University\\
{\small $^{*}$Corresponding Author, Contact: \href{mailto:rihan.zhang@duke.edu}{rihan.zhang@duke.edu}}}
\hypersetup{pdftitle={GRUN: Generalized Retinex Unfolding Network for Image Restoration},pdfauthor={Rihan Zhang, Fengyang Xiao, Yuhan Chen, Chunming He, Sina Farsiu}}
\makeatletter
\apptocmd{\@maketitle}{\par\noindent\input{sec/grun_teaser}\par\vspace{10pt}}{}{
  \PackageError{grun}{Cannot place first-page figure}{Check the CVPR title block.}}
\makeatother
\begin{document}
\raggedbottom
\maketitle
\input{sec/grun_abstract}

\input{sec/grun_body}

\par
{\small
\bibliographystyle{ieeenat_fullname}
\bibliography{main}
}
\clearpage
\end{document}

%% file: preamble.tex
\usepackage[T1]{fontenc}
\usepackage{amsthm,mathtools,mathrsfs,bm,bbm}
\usepackage{multirow,colortbl,makecell,tabularx}
\usepackage{algorithm,algorithmic}
\usepackage{microtype}
\usepackage{placeins}

\newcommand{\R}{\mathbf{R}}
\newcommand{\LL}{\mathbf{L}}
\newcommand{\x}{\mathbf{x}}
\newcommand{\y}{\mathbf{y}}
\newcommand{\z}{\mathbf{z}}

\newcommand{\Loss}{\mathcal{L}}
\newcommand{\thetadeg}{\bm{\theta}_{\mathrm{deg}}}
\newcommand{\Dc}{\mathcal{D}_c}
\newcommand{\De}{\mathcal{D}_e}

\definecolor{c2}{HTML}{FBD9BD}
\definecolor{c3}{HTML}{FE793D}
\definecolor{c4}{HTML}{EEDEB0}
\definecolor{pp}{HTML}{BC7FCD}
\definecolor{bb}{HTML}{CDE8E5}
\definecolor{rouse}{rgb}{0.981,0.961,0.941}
\makeatletter
\newcommand*\bigcdot{\mathpalette\bigcdot@{.5}}
\newcommand*\bigcdot@[2]{\mathbin{\vcenter{\hbox{\scalebox{#2}{$\m@th#1\bullet$}}}}}
\makeatother

\makeatletter
\newcommand{\GRUNRowBand}{%
  \leavevmode\rlap{\color{c2!20}%
    \rule[-\dp\@arstrutbox]{\linewidth}{\dimexpr\ht\@arstrutbox+\dp\@arstrutbox\relax}}}
\newcommand{\GRUNCellBand}[2][c2!20]{%
  \begingroup\setbox\z@\hbox{#2}%
  \leavevmode\rlap{\color{#1}%
    \rule[-\dp\@arstrutbox]{\dimexpr\wd\z@+18pt\relax}{\dimexpr\ht\@arstrutbox+\dp\@arstrutbox\relax}}%
  \box\z@\endgroup}
\makeatother

\input{sec/grun_layout}

%% file: sec/grun_layout.tex
\DeclareMathSizes{7.2}{7.2}{5}{5}
\DeclareMathSizes{6.5}{6.5}{5}{5}
\makeatletter
\let\GRUN@original@addtodblcol\@addtodblcol
\def\@addtodblcol{%
  \ifcsname GRUN@target@\number\@currbox\endcsname
    \ifnum\c@page<\csname GRUN@target@\number\@currbox\endcsname\relax
      \begingroup\@dbltopnum=-1\relax\GRUN@original@addtodblcol\endgroup
    \else\GRUN@original@addtodblcol\fi
  \else\GRUN@original@addtodblcol\fi}
\newenvironment{GRUNPageFloat}[1]{%
  \begin{figure*}[!t]
  \expandafter\xdef\csname GRUN@target@\number\@currbox\endcsname{#1}%
  \centering\setlength{\parskip}{0pt}%
}{\end{figure*}}
\makeatother

\AtBeginDocument{%
  \setlength{\abovedisplayskip}{2pt}%
  \setlength{\belowdisplayskip}{2pt}%
  \setlength{\abovedisplayshortskip}{2pt}%
  \setlength{\belowdisplayshortskip}{2pt}}
\newsavebox{\GRUNEquationBox}
\newcommand{\GRUNFitEquation}[1]{%
  \sbox{\GRUNEquationBox}{$\displaystyle #1$}%
  \ifdim\wd\GRUNEquationBox>\dimexpr\linewidth-3.4em\relax
    \resizebox{\dimexpr\linewidth-3.4em\relax}{!}{\usebox{\GRUNEquationBox}}%
  \else\usebox{\GRUNEquationBox}\fi}

%% file: sec/grun_teaser.tex
\begin{minipage}{\textwidth}
\normalfont\normalsize
\centering
\includegraphics[width=\linewidth]{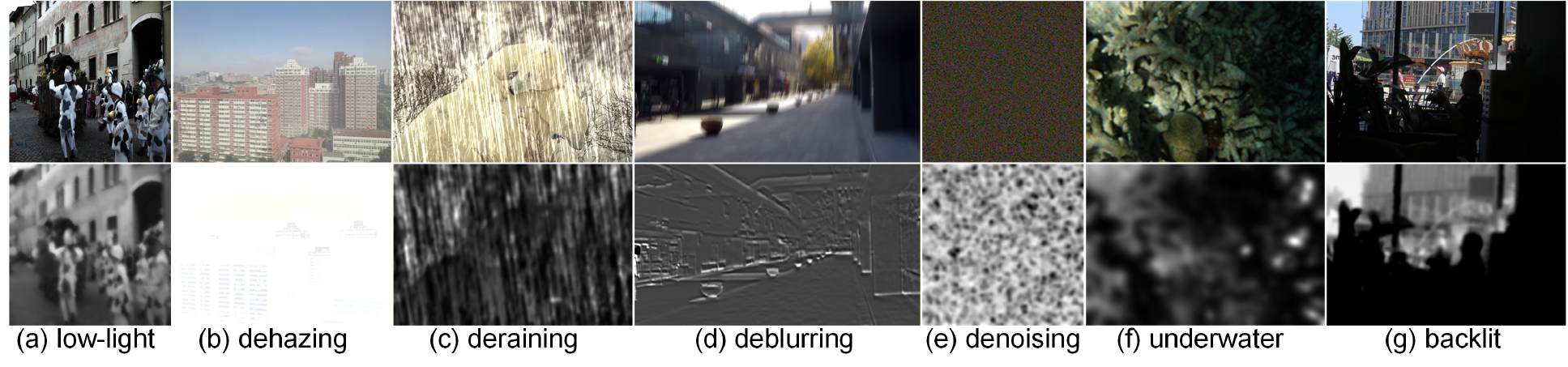}
\captionsetup{hypcap=false}\captionof{figure}{Natural images (\textbf{first row}) captured in uncontrolled scenes often suffer from diverse degradations such as noise, haze, rain, low illumination, and motion blur. The generalized illumination factor $L$ (\textbf{second row}) is visualized under our Generalized Retinex formulation, where $L$ denotes a task-dependent environment-side response.}\label{fig:motivation}
\end{minipage}

%% file: sec/grun_abstract.tex
\begin{abstract}
\label{abstract}
Image restoration remains a fundamental challenge in low-level vision, requiring methods that handle diverse degradations while preserving scene structures and details.
We revisit classical Retinex theory and propose a \textbf{Generalized Retinex} formulation that extends the illumination--reflectance decomposition into a Retinex-inspired content--environment factorization prior for image restoration. 
The key idea is to restore images using two learned factors: a \textbf{content-dominant response}
(generalized reflectance) capturing scene properties
and an \textbf{environment-side response}
(generalized illumination) encoding task-dependent variations associated with the observed degradations.
Based on this formulation, we develop \textbf{GRUN}
(\textbf{G}eneralized \textbf{R}etinex \textbf{U}nfolding
\textbf{N}etwork), a MAP-guided deep unfolding framework
that alternates reflectance restoration, illumination
estimation, and progressive degradation inference
across $K$ stages with degradation-conditioned
proximal operators.
We further introduce a \textbf{Degradation-aware
Coupling Regularizer} (DCR) that combines mutual
information minimization, gradient decorrelation,
and scale anchoring to promote complementary
content--environment factor representations.
Experiments on seven image restoration tasks, including denoising, dehazing, deraining, low-light enhancement, deblurring, underwater enhancement, and backlit enhancement, demonstrate competitive results on multiple benchmarks using the same architecture with task-specific weights.
Moreover, the proposed Generalized Retinex design
can be integrated into existing restoration methods,
demonstrating its cross-architecture applicability.
Code will be released.
\end{abstract}

%% file: sec/grun_body.tex
\begin{figure*}[t!]
  
  \centering
    \includegraphics[width=1\linewidth]{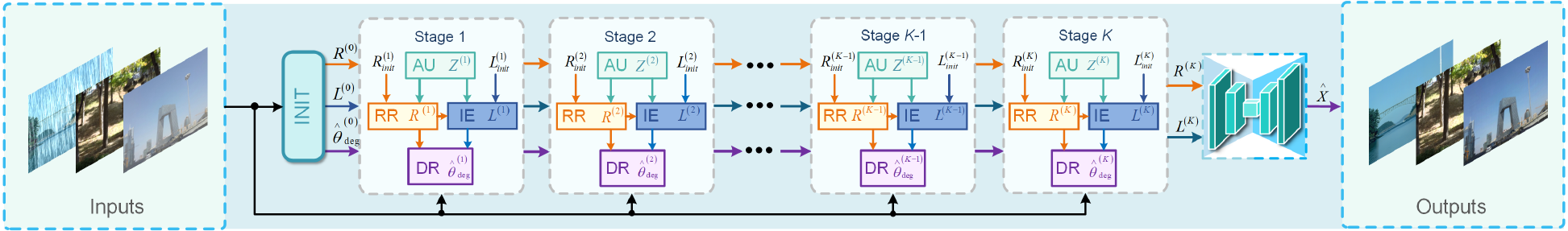}

    \caption{Framework of our proposed GRUN.}
    \label{Fig.framework_all}
\vspace{-6mm}
\end{figure*}

\section{Introduction} \label{introduction} 
Natural images captured in uncontrolled environments often suffer from degradations such as noise, haze, rain, low illumination, and motion blur. Restoring clean images from such observations is a long-standing problem in low-level vision.
Image restoration methods often rely on degradation-specific models and priors, such as atmospheric scattering models for dehazing, Retinex-based illumination modeling for low-light enhancement~\cite{wei2018deep}, and point spread functions for deblurring. While physically motivated, these formulations rely on assumptions tailored to specific degradation mechanisms, limiting their direct applicability across restoration tasks.
Deep unfolding bridges model-based optimization and deep learning by unrolling iterative algorithms into trainable networks. Existing methods~\cite{he2025unfoldir,mou2022deep,deng2025deepsn} have achieved strong performance across diverse restoration tasks through flexible optimization frameworks and learned priors. Nevertheless, effectively modeling the interactions between underlying scene structures and diverse degradation effects remains an important challenge in image restoration.

We observe that the classical Retinex theory~\cite{land1977retinex}---originally formulated for illumination modeling---encapsulates a more general content--environment separation principle that extends beyond low-light enhancement. The Retinex decomposition $I = R \odot L$ separates an observed image into a reflectance component $R$ and an illumination component $L$. We generalize this idea by treating $R$ as a content-dominant response and $L$ as a task-dependent environment-side response. 
As illustrated in Fig.~\ref{fig:motivation},
despite diverse degradation patterns across
restoration tasks, the generalized illumination
factor $L$ captures task-dependent environment-side variations associated with the observed degradations.
We argue that this ``\emph{content vs.\ environment}'' factorization provides a powerful structural prior for diverse image degradations. 
In low-light enhancement, $L$ models illumination-
or exposure-related variations, while $R$ preserves intrinsic scene colors and textures.
In image dehazing, $L$ captures haze-related
environment-side variations associated with
atmospheric scattering, while $R$ captures scene structures and details obscured by haze.
In deraining, $R$ captures background structures and textures obscured by rain streaks, while $L$ summarizes rain-induced occlusion, visibility attenuation, or spatial modulation.
For deblurring, $L$ characterizes spatially varying blur-related modulation or degradation strength, while $R$ captures sharp scene structures and fine details.
In denoising, $R$ captures the underlying image structures and details, while $L$ serves as an identity-like environment-side response that can encode spatially varying uncertainty.
Similarly, in underwater and backlit enhancement, $L$ captures environment-side variations associated with medium-dependent color changes, illumination shifts, or shadow modulation, while $R$ preserves the intrinsic scene appearance.
This observation motivates us to formulate a Generalized Retinex decomposition as a Retinex-inspired content--environment factorization prior for image restoration.
Building on this formulation, we develop GRUN, a MAP-guided deep unfolding framework that alternates reflectance restoration, illumination estimation, and progressive degradation inference.
We further introduce a Degradation-aware Coupling Regularizer (DCR) that combines mutual information minimization, gradient decorrelation, and scale anchoring to encourage complementary content--environment representations. Following prior unfolding methods~\cite{mou2022deep,he2025unfoldldm,deng2025deepsn}, GRUN adopts the same architecture with task-specific weights across tasks.
We make the following contributions:

\noindent(1) We propose a \textbf{Generalized Retinex formulation} that generalizes the classical illumination--reflectance decomposition into a Retinex-inspired content--environment factorization prior for image restoration, where $\R$ represents a content-dominant response capturing scene properties and $\LL$ denotes an environment-side response encoding task-dependent variations associated with the degradations.

\noindent(2) We develop \textbf{GRUN}, a MAP-guided deep unfolding network based on the generalized Retinex model, featuring alternating reflectance restoration, illumination estimation, and progressive degradation inference with degradation-conditioned proximal operators.

\noindent(3) We introduce a \textbf{Degradation-aware Coupling Regularizer (DCR)} that combines mutual information minimization, gradient decorrelation, and scale anchoring to promote complementary content--environment factor representations.

\noindent(4) Extensive experiments on seven image restoration tasks demonstrate competitive results on multiple benchmarks with task-adaptive intermediate decompositions, while integration into existing restoration architectures further demonstrates the applicability of our approach.

\section{Related Work}
\textbf{Deep Unfolding-based Image Restoration.} Deep unfolding networks (DUNs)~\cite{he2023degradation} unroll iterative optimization into trainable architectures and have been applied to deblurring~\cite{mou2022deep}, super-resolution~\cite{zhang2020deep}, and low-light enhancement~\cite{wu2022uretinex,he2025unfoldir}. Representative methods include DGUNet~\cite{mou2022deep}, DeepSN-Net~\cite{deng2025deepsn}, and UnfoldLDM~\cite{he2025unfoldldm}. 
Nevertheless, incorporating interpretable structural guidance into the unfolding process remains an important direction for image restoration.

\noindent\textbf{Retinex-based Methods.}
Retinex theory~\cite{land1977retinex} decomposes images into reflectance and illumination and has been widely used for low-light enhancement. RetinexNet~\cite{wei2018deep}, KinD~\cite{zhang2019kindling}, URetinex-Net~\cite{wu2022uretinex}, Diff-Retinex~\cite{yi2023diff}, and Reti-Diff~\cite{he2023reti} improve decomposition using deep, unfolding, or diffusion models. Yet, these methods mainly focus on illumination-related enhancement and rarely generalize the Retinex prior to broader degradation types.

\noindent\textbf{Prior-guided Image Restoration.}
Image priors play a crucial role in constraining
ill-posed restoration problems.
Handcrafted priors such as total variation~\cite{chambolle2004algorithm}
impose explicit regularity, while learned generative
priors, including diffusion models~\cite{xia2023diffir,he2023reti},
capture complex image statistics.
In contrast, we introduce a Retinex-inspired
content--environment factorization prior that
provides structured guidance for restoration
within a deep unfolding framework.

\begin{figure*}[t!]
  
  \centering
    \includegraphics[width=1\linewidth]{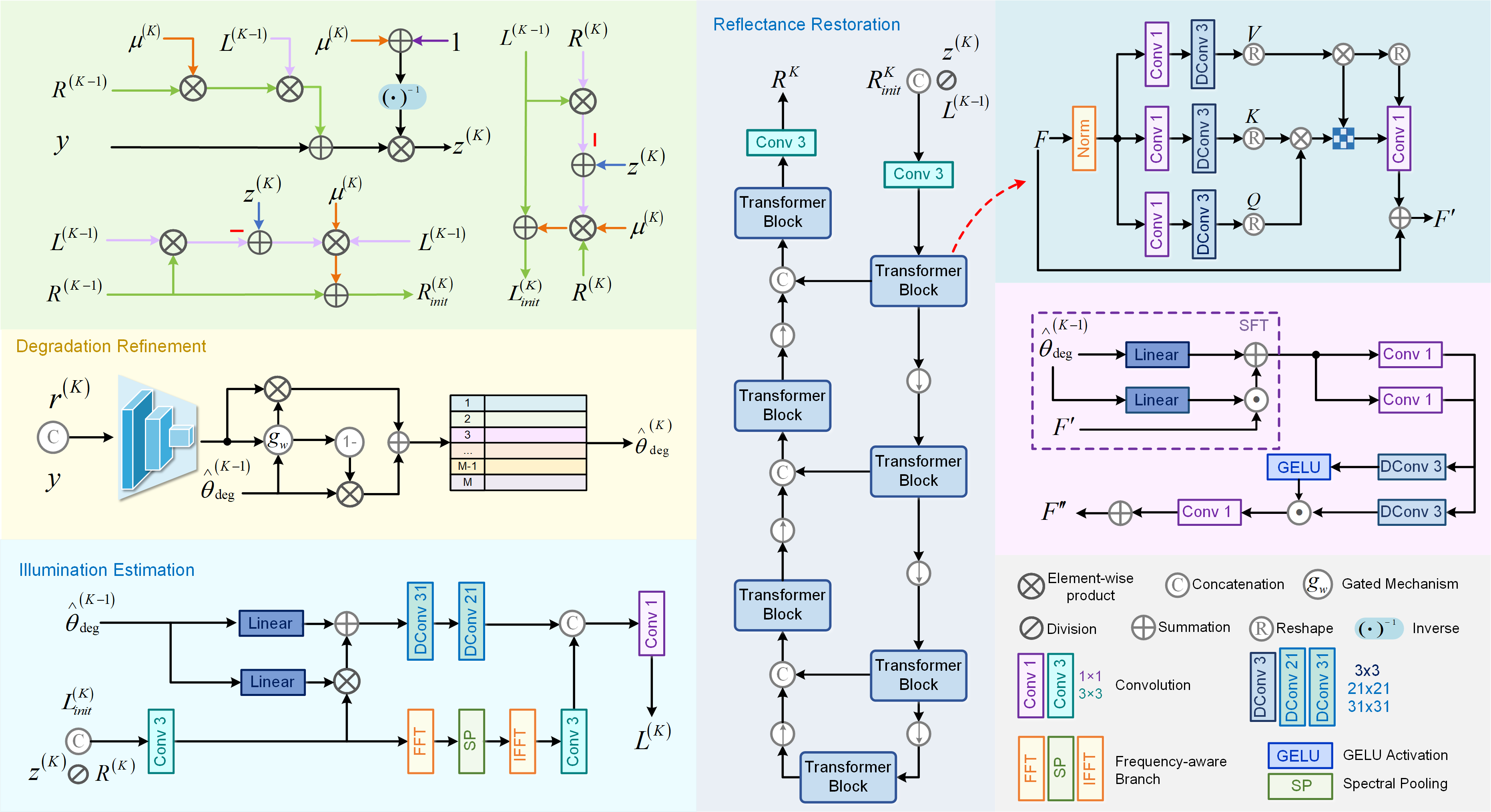}

    \caption{Details of the $k$-th stage in the GRUN unfolding framework.}
    \label{Fig.framework}
    \vspace{-6mm}

\end{figure*}
\section{Methodology}
\label{sec:method}
As shown in Fig.~\ref{Fig.framework_all} and Fig.~\ref{Fig.framework}, we propose GRUN, a deep unfolding framework built upon the generalized Retinex formulation. 
The network iteratively updates the reflectance and illumination factors while progressively refining the degradation representation across multiple stages.
\subsection{Generalized Retinex Formulation}
\label{sec:gen_retinex}
The classical Retinex theory models an observed image as the element-wise product of reflectance and illumination: $\mathbf{I} = \R \odot \LL$. 
We generalize this principle by introducing task-dependent degradation operators for both factors, yielding the \textbf{Generalized Retinex formulation}:
\begin{equation}
\label{eq:gen_retinex}
\GRUNFitEquation{\y = \De(\LL) \odot \Dc(\R) + \mathbf{n},}
\end{equation}
where $\R \in \mathbb{R}^{H \times W \times C}$ denotes the \textbf{content-dominant response} (generalized reflectance), $\LL \in \mathbb{R}^{H \times W \times C}$ denotes the \textbf{environment-side response} (generalized illumination). Here, $\Dc(\cdot)$ and $\De(\cdot)$ denote task-dependent content-side and environment-side operators, respectively. 
The former can model content-side distortions associated with noise, rain streaks, or blur, while the latter can characterize environment-side modulation related to illumination variation, transmission/attenuation, occlusion, uncertainty, or spatially varying degradation strength.
The term $\mathbf{n}$ denotes additive residual noise.
\cref{eq:gen_retinex} lifts classical Retinex from an illumination--reflectance image formation model to a content--environment factorization for diverse degradations.
Under this formulation, $\R$ represents a content-dominant response, while $\LL$ encodes task-dependent environment-side variations associated with the observed degradations. The task-dependent operator mappings provide degradation-motivated interpretations rather than exact algebraic reformulations of the underlying physical models.
See Sec.~\ref{sec:supp_retinex_theory} of the supplementary material for detailed degradation mappings.
Given the degraded observation $\y$, our objective
is to estimate the latent generalized response factors
$\R^*$ and $\LL^*$, whose product provides a
factorized restoration estimate, while encouraging
complementary content-dominant and environment-side
representations.

\subsection{MAP-Guided Formulation}
\label{sec:map}
Motivated by the Generalized Retinex formulation in \cref{eq:gen_retinex}, we formulate the following MAP-guided optimization objective over the latent generalized factors $\R$ and $\LL$:
\begin{equation}
\label{eq:map}
\begin{aligned}
\min_{\R,\LL}\;&\frac{1}{2}\left\|\y-\De(\LL)\odot\Dc(\R)\right\|_2^2 \\
&{}+\lambda_R\Omega_R(\R)+\lambda_L\Omega_L(\LL)+\lambda_\Psi\Psi(\R,\LL),
\end{aligned}
\end{equation}
where $\Omega_R(\R)$, $\Omega_L(\LL)$, and
$\Psi(\R,\LL)$ encode factor-specific priors
and interactions.
The quadratic data-fidelity term promotes
consistency with the generalized observation model.

\textbf{Reflectance Prior} $\Omega_R(\R)$. The content-dominant response should exhibit piecewise smoothness with preserved edges and textures:
\begin{equation}
\GRUNFitEquation{\Omega_R(\R) = \|\nabla \R\|_{\text{ani}} + \gamma_R \|\R\|_{\text{NL}},}
\end{equation}
where $\|\cdot\|_{\text{ani}}$ denotes anisotropic total variation promoting edge-preserving smoothness, while $\|\cdot\|_{\text{NL}}$ represents a non-local self-similarity prior that motivates the structural design of the learned reflectance proximal operator.

\textbf{Illumination Prior} $\Omega_L(\LL)$. The environment-side response is encouraged to be spatially coherent and low-frequency dominant:
\begin{equation}
\GRUNFitEquation{\Omega_L(\LL) = \|\nabla \LL\|_2^2 + \gamma_L \|\LL - \bar{\LL}\|_{\mathcal{F}}^2,}
\end{equation}
where $\bar{\LL}$ denotes a low-pass version of $\LL$, and the frequency-domain term penalizes high-frequency deviations.
These priors motivate the large-kernel and frequency-aware design of the subsequent  learned illumination proximal operator.

\textbf{Coupling Regularizer} $\Psi(\R, \LL)$. 
To reduce factor redundancy and encourage
complementary representations: 
\begin{equation}
\label{eq:coupling}
\begin{aligned}
\Psi(\R,\LL) &= \widehat{\text{MI}}(\R,\LL)
+\eta\|\text{corr}(\nabla\R,\nabla\LL)\|_F^2 \\
&\quad +\zeta(\text{mean}(\LL)-1)^2,
\end{aligned}
\end{equation}
where $\widehat{\text{MI}}(\cdot,\cdot)$ denotes the mutual information, the gradient correlation term discourages duplicated edge structures, and the scale term mitigates the scale ambiguity of the Retinex.

\subsection{Deep Unfolding via Half-Quadratic Splitting}
\label{sec:unfolding}
To develop a tractable unfolding framework from \cref{eq:map}, we employ half-quadratic splitting (HQS) to decouple observation fidelity from the generalized factor coupling.
Direct optimization of the resulting subproblems would require the unknown task-dependent operators and their Jacobians.
We therefore adopt a tractable multiplicative surrogate for explicit factor updates.
To account for degradation effects not explicitly captured by these updates, the progressive degradation estimator refines degradation embeddings across stages, which condition the proximal operators for task-adaptive factor refinement.
Meanwhile, DCR mitigates factor redundancy and scale ambiguity by encouraging complementary content--environment representations during training.
Specifically, we introduce an auxiliary variable associated with the surrogate coupling and unroll $K$ alternating stages to update $\z$, $\R$, $\LL$, and the degradation embedding.

At stage $k$, the auxiliary variable admits
a closed-form update under the surrogate formulation:
\begin{equation}
\label{eq:z_update}
\GRUNFitEquation{\z^{(k)}
=
\frac{\y+\mu^{(k)}\R^{(k-1)}\odot\LL^{(k-1)}}{1+\mu^{(k)}}.}
\end{equation}
This update combines the observation $\y$ with the previous factorized estimate $\R^{(k-1)}\odot\LL^{(k-1)}$, with the learnable parameter $\mu^{(k)}$ controlling their relative contributions.

For the $\R$- and $\LL$-subproblems, we take
one gradient descent step on the surrogate
data-fidelity term to obtain the proximal inputs:
\begin{gather}
\GRUNFitEquation{\R_{\mathrm{init}}^{(k)}
=
\R^{(k-1)}
+
\mu^{(k)}\LL^{(k-1)}
\odot
\left(
\z^{(k)}
-
\R^{(k-1)}\odot\LL^{(k-1)}
\right),}\\
\GRUNFitEquation{\LL_{\mathrm{init}}^{(k)}
=
\LL^{(k-1)}
+
\mu^{(k)}\R^{(k)}
\odot
\left(
\z^{(k)}
-
\R^{(k)}\odot\LL^{(k-1)}
\right).}
\end{gather}
The gradient-updated proximal inputs are then refined by degradation-conditioned proximal operators:
\begin{equation}
\label{eq:prox_updates}
\GRUNFitEquation{\begin{aligned}
\R^{(k)} = \mathcal{P}_R^{(k)}\!\left(
\R_{\mathrm{init}}^{(k)},\,
\z^{(k)}\oslash\LL^{(k-1)},\,
\hat{\thetadeg}^{(k-1)}
\right),  \\
\LL^{(k)} = \mathcal{P}_L^{(k)}\!\left(
\LL_{\mathrm{init}}^{(k)},\,
\z^{(k)}\oslash\R^{(k)},\,
\hat{\thetadeg}^{(k-1)}
\right),  \\
\hat{\thetadeg}^{(k)} = \mathcal{E}^{(k)}\!\left(
\y,\,
\R^{(k)},\,
\LL^{(k)},\,
\hat{\thetadeg}^{(k-1)}
\right).
\end{aligned}}
\end{equation}
Here $\oslash$ denotes numerically stabilized element-wise division. 
The three learnable update modules, namely, $\mathcal{P}_R^{(k)}$, $\mathcal{P}_L^{(k)}$, and $\mathcal{E}^{(k)}$, instantiate the proximal and estimation steps of the unfolded optimization with task-adaptive neural networks. 
They use different inductive biases for content recovery, environment-side response estimation, and degradation inference, respectively: $\mathcal{P}_R^{(k)}$ preserves structures and textures, $\mathcal{P}_L^{(k)}$ favors spatially coherent degradation responses, and $\mathcal{E}^{(k)}$ refines the degradation embedding across stages.
Details are provided in~\cref{sec:proximal,sec:deg_est}. 
The full derivation from the HQS subproblems is provided in Sec.~\ref{sec:app_hqs_derivation}.

\paragraph{Initialization.}
Before unfolding, a lightweight initialization module estimates $(\R^{(0)},\LL^{(0)})$ and the initial degradation embedding $\hat{\thetadeg}^{(0)}$ from the degraded input $\y$. We initialize $L^{(0)}$ using a sigmoid-based parameterization, while detailed architecture and initialization equations are provided in Sec.~\ref{sec:app_init_details}. The complete stage-wise procedure is summarized in Algorithm~\ref{alg:grun}.

\subsection{Degradation-Conditioned Proximal Operators}
\label{sec:proximal}

The learnable proximal operators $\mathcal{P}_R^{(k)}$ and $\mathcal{P}_L^{(k)}$ are the computational core of each unfolding stage. Each operator uses a factor-specific architecture with its own inductive bias.

\textbf{Reflectance Proximal Operator} $\mathcal{P}_R^{(k)}$. 
Since the generalized reflectance $\R$ primarily captures scene structures and fine details, we employ a multi-scale encoder-decoder with \textbf{Restormer-style} channel self-attention blocks~\cite{zamir2022restormer}. The architecture takes three inputs: (i) the gradient-initialized reflectance $\R_{\text{init}}^{(k)}$, (ii) the quotient estimate $\z^{(k)} \oslash \LL^{(k-1)}$ as a complementary cue, and (iii) the degradation embedding $\hat{\thetadeg}^{(k-1)}$. The degradation embedding is injected via \textbf{Spatial Feature Transform (SFT)}~\cite{wang2018sft}:
\begin{equation}
\GRUNFitEquation{\text{SFT}(\mathbf{F}, \hat{\thetadeg}) = \bm{\gamma}(\hat{\thetadeg}) \odot \mathbf{F} + \bm{\beta}(\hat{\thetadeg}),}
\end{equation}
where $\bm{\gamma}$ and $\bm{\beta}$ are learned affine parameters generated from $\hat{\thetadeg}$ through two linear layers, and $\mathbf{F}$ denotes intermediate features. This conditions feature scaling and shifting on the learned
embedding, enabling task-specific refinement at each stage.

\textbf{Illumination Proximal Operator} $\mathcal{P}_L^{(k)}$. 
The generalized illumination factor $\LL$ encodes environment-side variations and is encouraged to exhibit spatial coherence.
We therefore adopt a lightweight network with large-kernel depthwise convolutions ($31\times31$ and $21\times21$) and a frequency-aware low-frequency branch via spectral pooling. 
The degradation embedding $\hat{\thetadeg}^{(k-1)}$ is injected through SFT modulation, as in $\mathcal{P}_R^{(k)}$. 
After the final stage, the estimated generalized reflectance and illumination are fused by a lightweight reconstruction network~\cite{he2025unfoldir} to produce $\hat{\x}=F_{\theta}(\R^{(K)},\LL^{(K)})$.

\input{sec/grun_page6}
\subsection{Progressive Degradation Estimator}
\label{sec:deg_est}

The degradation estimator $\mathcal{E}^{(k)}$ produces a continuous degradation embedding $\hat{\thetadeg}^{(k)} \in \mathbb{R}^d$ for task-specific restoration,
without explicit degradation-type classification.

\textbf{Architecture.} 
At each stage $k$, the estimator computes the
residual $\mathbf{r}^{(k)}=\y-\R^{(k)}\odot\LL^{(k)}$
between the observation and the current factorized
estimate. A lightweight convolutional encoder extracts
multi-scale features from the residual and observation,
which are fused with the previous degradation embedding
$\hat{\thetadeg}^{(k-1)}$ via a Gated Recurrent Unit (GRU)~\cite{cho2014gru}:
\begin{equation}
\GRUNFitEquation{\hat{\thetadeg}^{(k)} = \text{GRU}\!\left(\text{Enc}(\mathbf{r}^{(k)}, \y),\; \hat{\thetadeg}^{(k-1)}\right).}
\end{equation}

\textbf{Degradation Codebook.} To structure the degradation embedding space, we maintain a learnable codebook $\mathcal{C} = \{\mathbf{c}_1, \ldots, \mathbf{c}_M\} \subset \mathbb{R}^d$,
whose code vectors serve as prototypes in the learned
conditioning space. The final embedding is computed via soft attention:
\begin{equation}
\GRUNFitEquation{
\hat{\thetadeg}^{(k)}_{\text{final}}
=\sum_{m=1}^{M}\alpha_m^{(k)}\mathbf{c}_m,  
\alpha_m^{(k)}
=\frac{\exp(\hat{\thetadeg}^{(k)}\cdot\mathbf{c}_m/\tau)}
{\sum_{j=1}^{M}\exp(\hat{\thetadeg}^{(k)}\cdot\mathbf{c}_j/\tau)},
}
\end{equation}
where $\tau$ is a temperature parameter. With task-specific training, the soft prototype mixture compactly encodes intra-task degradation variation to condition the proximal operators.

\subsection{Degradation-aware Coupling Regularizer }
\label{sec:dcr}

The coupling regularizer in~\cref{eq:coupling}
helps mitigate degenerate decompositions.
We implement it through three complementary loss terms.

\textbf{Mutual Information Minimization.} We use the MINE estimator~\cite{belghazi2018mutual} to approximate the mutual information between $\R^{(k)}$ and $\LL^{(k)}$ at each stage. A small critic network $T_\phi$ parameterizes the Donsker-Varadhan bound:
\begin{equation}
\GRUNFitEquation{
\Loss_{\text{MI}}^{(k)}=\sup_\phi\Bigl[\mathbb{E}_{p(\R^{(k)},\LL^{(k)})}[T_\phi]  
-\log\mathbb{E}_{p(\R^{(k)})p(\LL^{(k)})}[e^{T_\phi}]\Bigr].
}
\end{equation}
In training, the critic network $T_\phi$ is optimized to maximize the MINE objective, while the restoration network is optimized to minimize the estimated mutual information via alternating gradient updates.

\textbf{Gradient Decorrelation.} We penalize the cosine similarity between gradient maps:
\begin{equation}
\GRUNFitEquation{\Loss_{\text{grad}}^{(k)} = \frac{|\langle \nabla \R^{(k)}, \nabla \LL^{(k)} \rangle|}{\|\nabla \R^{(k)}\|_F \cdot \|\nabla \LL^{(k)}\|_F + \epsilon}.}
\end{equation}
This helps discourage duplication of edge structures
across both factors, a potential issue in Retinex decomposition.

\textbf{Scale anchoring.} 
The multiplicative factorization admits reciprocal
scale ambiguity:
$(a\R)\odot(\LL/a)=\R\odot\LL$ for any $a>0$.
To mitigate this ambiguity, we penalize deviations
of the spatial mean of $\LL$ from unity:
\begin{equation}
\GRUNFitEquation{\Loss_{\text{scale}}^{(k)} = \left(\text{mean}(\LL^{(k)}) - 1\right)^2,}
\end{equation}
which anchors $\LL$ around unity and stabilizes the factorized representation.

\subsection{Training Objective}
\label{sec:training}

GRUN is trained end-to-end with deep supervision over all unfolding stages. The overall objective is
\begin{equation}
\label{eq:total_loss}
\GRUNFitEquation{\Loss_{\text{total}}
=
\Loss_{\text{recon}}
+
\alpha \Loss_{\text{decomp}}
+
\beta \Loss_{\text{deg}}
+
\gamma \Loss_{\text{DCR}},}
\end{equation}
where $\Loss_{\text{recon}}$ supervises the restored image, $\Loss_{\text{decomp}}$ regularizes the generalized Retinex response factors, $\Loss_{\text{deg}}$ structures the degradation embedding space, and $\Loss_{\text{DCR}}$ encourages complementary content--environment
factor representations. Detailed definitions of all loss terms and weighting coefficients are provided in Sec.~\ref{sec:supp_training_objective}.

\input{sec/grun_page7}
\section{Experiments}
\label{sec:Experiments}
\noindent \textbf{Experimental setup}.
We use $K=3$ unfolding stages and train task-specific weights while keeping the same GRUN architecture across all tasks. 
The degradation embedding dimension is set to $d=256$, with codebook size $M=64$, temperature $\tau=0.10$, and a 512-dim GRU-based estimator. 
MINE is used only during training to regularize the decomposition, with warm-up and stabilization details provided in Sec.~\ref{sec:supp_training_objective}. 
All models are trained from scratch using AdamW with an initial learning rate of $2\times10^{-4}$, cosine annealing, batch size 8, and $256\times256$ image patches on one NVIDIA H200 GPU.
For fair comparison, we follow the standard training/testing protocols and evaluation metrics of each benchmark.
Datasets are detailed in Sec.~\ref{sec:supp_datasets}.

\subsection{Comparative Evaluation: Paired Benchmarks}
We evaluate tasks using PSNR and SSIM, covering single-type degradations
(denoising, deblurring, deraining, dehazing) and compound degradations
(low-light, underwater, backlit enhancement).

\noindent\textbf{Denoising and Deblurring.}
Following DeepSN-Net~\cite{deng2025deepsn}, we train on SIDD~\cite{abdelhamed2018high} and evaluate on SIDD and DND~\cite{plotz2017benchmarking}.
For deblurring, we train on GoPro~\cite{nah2017deep} and test on GoPro and HIDE~\cite{shen2019human}, following~\cite{he2025unfoldldm,deng2025deepsn}.
As shown in \cref{table:denoising,table:deblur}, GRUN achieves the best quantitative performance among the compared methods.
Visual results in \cref{Fig.denoise,Fig.deblur} show finer details and sharper structures.

\noindent\textbf{Deraining and Dehazing.}
Following~\cite{deng2025deepsn,song2023vision}, we evaluate on Rain100H/L~\cite{yang2017deep}, Test100~\cite{zhang2019image}, Test2800~\cite{fu2017removing}, Test1200~\cite{zhang2018density}, and RESIDE~\cite{li2018benchmarking}.
As shown in \cref{table:deraining,table:dehazing}, GRUN achieves strong quantitative performance across these benchmarks.
Visual results in \cref{Fig.LOL_backlit_underwater} show effective rain removal and natural dehazing.

\noindent\textbf{Low-Light, Underwater, and Backlit Enhancement.}
Following~\cite{he2023reti,he2025unfoldir}, we evaluate on LOL-v1~\cite{wei2018deep}, LOL-v2-real/syn~\cite{yang2021sparse}, and UIEB~\cite{li2019underwater}.
For backlit enhancement, we train and evaluate on BAID~\cite{lv2022backlitnet}, following~\cite{liang2023iterative}.
Results in \cref{table:low-light,table:Underwater,table:backlit} demonstrate effective restoration across these tasks.
Visual results in \cref{Fig.LOL_backlit_underwater} show natural brightness, consistent colors, and clearer details.

\input{sec/grun_page8}
\subsection{Comparative Evaluation on Complex Real-world Degradations}
\noindent \textbf{Real-world IDIR}. Three real-world illumination degradation image restoration (IDIR) datasets, DICM~\cite{lee2013contrast}, VV~\cite{vonikakis_vv}, and MEF~\cite{ma2015perceptual} test no-reference generalization of a model pretrained on \textit{LOL-v2-syn}. As shown in \cref{table:realIDIR}, GRUN achieves the lowest PI and NIQE on all three datasets.

\subsection{Ablation Study}
\noindent\textbf{Effectiveness of Key GRUN Components.}
We ablate Generalized Retinex Decomposition (GRD), degradation estimation (DE), and DCR.
For GRD, we remove $R^{(k)}$, $L^{(k)}$, $z^{(k)}$, the reconstruction head $F_{\theta}$, or the entire Retinex formulation.
We also replace the factor-specific proximal operators $\mathcal{P}_L^{(k)}$ and $\mathcal{P}_R^{(k)}$ with standard convolutional networks, denoted by $\mathcal{L}_1(\cdot)$ and $\mathcal{L}_2(\cdot)$, respectively.
For DE, we remove the estimator or codebook, or replace the estimator with a CNN ($\mathcal{D}_1$) or VSSBlock ($\mathcal{D}_2$).
All variants reduce restoration quality (\cref{table:Ablation}); removing GRD, DE, or DCR also causes color and contrast distortions (\cref{Fig.ablation}).
Compared with GRUN without DCR, full GRUN shows lower estimated mutual
information and lower gradient correlation between $R$ and $L$ on all
five tasks in \cref{tab:coupling}.
Cross-task ablations in Sec.~\ref{sec:supp_ablation}   (\cref{table:cross_task_ablation}) demonstrate
the consistent contributions of GRUN components
across different restoration tasks.
Under the same branch architectures and degradation conditioning,
replacing the explicit updates and quotient cues with learned fusion yields Generic Two-Branch.
This baseline is trained without $\Loss_{\text{decomp}}$ and the entire $\Loss_{\text{DCR}}$ term.
As shown in \cref{tab:retinex_pathway}, full GRUN achieves higher PSNR and SSIM
than this baseline on both GoPro and SIDD.

\noindent\textbf{Effectiveness of Generalized Retinex Factorization.}
We evaluate multiplicative coupling and cross-factor interactions against task-specific alternatives across five restoration tasks.
As shown in \cref{tab:coupling}, the full product--quotient formulation (P+Q) consistently outperforms both task-specific coupling with matched cues (T+M) and product coupling with generic cues (P+G), validating the effectiveness of generalized Retinex factorization and quotient-based interactions.
Detailed coupling formulations and implementation protocols are provided in Sec.~\ref{sec:supp_further_analysis}.

\subsection{Further Analysis}

\noindent\textbf{Interpretability of Generalized Retinex Factors.}
We investigate the functional roles of generalized reflectance $R$ and illumination $L$.
As visualized in \cref{fig:decomposition}, $R$ preserves scene structures, while $L$ captures task-dependent environment-side responses.
Quantitative factor--proxy analyses across seven restoration tasks further reveal strong clean-structure correlations for $R$ and task-dependent degradation alignment for $L$.
Complete results and metric definitions are provided in Sec.~\ref{sec:supp_further_analysis} and \cref{tab:factor_roles_full} of the supplementary material.

\noindent\textbf{Cross-Architecture Applicability.}
We integrate GRUN components into six restoration architectures across six tasks.
For architectures without an explicit Retinex decomposition, we introduce a GRUN-style decomposition head, degradation embedding, and DCR loss. 
As shown in \cref{table:generalization}, these components improve DGUNet on deblurring, DeepSN-Net on deraining, and CORUN on dehazing.
For architectures with an existing Retinex decomposition, we preserve their original decomposition structures and incorporate the degradation embedding and DCR.
The adapted UnfoldIR, Reti-Diff, and URetinex-Net also improve on underwater, backlit, and low-light enhancement, respectively.

\noindent\textbf{Degradation Codebook Analysis.}
Increasing the codebook size from 8 to 64 consistently improves restoration performance on both LOL-v2 splits, while 128 brings no further gains.
Across seven tasks, prototype utilization on held-out patches is broad without evident collapse (\cref{table:stage_codebook_mine_ablation,tab:codebook_utilization}).

\noindent \textbf{Benefits for downstream applications}. 
Following~\cite{he2023reti}, we retrain YOLO on \textit{ExDark}~\cite{loh2019getting} enhanced by each method. GRUN achieves the highest mean AP, exceeding the strongest enhancement baseline by 1.2 points (\cref{table:Detection}).

\noindent\textbf{Cross-Stage Evolution and Efficiency.}
We analyze the progressive restoration process across unfolding stages.
As illustrated in \cref{fig:cross_stage}, 
GRUN progressively refines image structures and details. Sec.~\ref{sec:supp_further_analysis} shows that increasing $K$ improves restoration quality at additional computational cost (\cref{table:stage_codebook_mine_ablation,tab:efficiency_2}).
With $K=3$, GRUN requires 21.54M parameters, 83.87G FLOPs, and 40.37 ms per image at $256\times256$, achieving a favorable balance between restoration quality and efficiency on both LOL-v2 splits in the stage ablation.

\noindent\textbf{Robustness of MINE Training.}
GRUN maintains robust restoration performance across different MINE training strategies, without noticeable performance collapse.
While critic warm-up followed by gradual MI-weight ramping achieves the best results, the other strategies also remain effective on both LOL-v2 splits.
Detailed comparisons of the three schedules are provided in Sec.~\ref{sec:supp_ablation} (\cref{table:stage_codebook_mine_ablation}).

\section{Conclusions}
In this paper, we propose GRUN, a Generalized Retinex-based deep unfolding framework for image restoration.
By generalizing classical Retinex decomposition into a content--environment factorization, GRUN provides a unified structural prior that separates content-dominant responses from task-dependent environment-side responses.
Experiments across seven restoration tasks demonstrate
competitive performance using the same architecture
with task-specific weights.
Integrating GRUN components into six existing restoration
architectures also improves their reported results.

%% file: sec/grun_page6.tex
\begin{GRUNPageFloat}{6}

\noindent
  \begin{minipage}[t]{\linewidth}\vspace{0pt}\centering
  \captionsetup*{type=figure,position=bottom,skip=0.5pt,hypcap=false}%
  \setcounter{figure}{\numexpr4-1\relax}%
  \input{sec/layout_cells/main_figure_4.tex}\par
  \caption{Visualization of image denoising. Our method restores a more accurate ``i'' (right) and a ``p'' without trailing artifacts (left).}\label{Fig.denoise}%
  \end{minipage}\par
\vspace{2pt}

\noindent
  \begin{minipage}[t]{\linewidth}\vspace{0pt}\centering
  \captionsetup*{type=figure,position=bottom,skip=0.5pt,hypcap=false}%
  \setcounter{figure}{\numexpr5-1\relax}%
  \input{sec/layout_cells/main_figure_5.tex}\par
  \caption{Visualization of image deblurring. GRUN restores sharper hand details (left) and a
clearer ``D'' (right).}\label{Fig.deblur}%
  \end{minipage}\par
\vspace{2pt}

\noindent
  \begin{minipage}[t]{.40\linewidth}\vspace{0pt}\centering
  \captionsetup*{type=table,position=top,skip=0pt}%
  \setcounter{table}{\numexpr1-1\relax}%
  \begin{minipage}[t][11pt][t]{\linewidth}
    \caption{Results on denoising.}\label{table:denoising}%
  \end{minipage}\par\vspace{0pt}%
  \input{sec/layout_cells/main_table_1.tex}  \end{minipage}
\hfill
  \begin{minipage}[t]{.29\linewidth}\vspace{0pt}\centering
  \captionsetup*{type=table,position=top,skip=0pt}%
  \setcounter{table}{\numexpr2-1\relax}%
  \begin{minipage}[t][11pt][t]{\linewidth}
    \caption{Results on the UIE task.}\label{table:Underwater}%
  \end{minipage}\par\vspace{0pt}%
  \input{sec/layout_cells/main_table_2.tex}  \end{minipage}
\hfill
  \begin{minipage}[t]{.29\linewidth}\vspace{0pt}\centering
  \captionsetup*{type=table,position=top,skip=0pt}%
  \setcounter{table}{\numexpr3-1\relax}%
  \begin{minipage}[t][11pt][t]{\linewidth}
    \caption{Results on the BIE task.}\label{table:backlit}%
  \end{minipage}\par\vspace{0pt}%
  \input{sec/layout_cells/main_table_3.tex}  \end{minipage}\par
\vspace{2pt}

\noindent
  \begin{minipage}[t]{.40\linewidth}\vspace{0pt}\centering
  \captionsetup*{type=table,position=top,skip=0pt}%
  \setcounter{table}{\numexpr4-1\relax}%
  \begin{minipage}[t][11pt][t]{\linewidth}
    \caption{Results on deblurring.}\label{table:deblur}%
  \end{minipage}\par\vspace{0pt}%
  \input{sec/layout_cells/main_table_4.tex}  \end{minipage}
\hfill
  \begin{minipage}[t]{.59\linewidth}\vspace{0pt}\centering
  \captionsetup*{type=table,position=top,skip=0pt}%
  \setcounter{table}{\numexpr5-1\relax}%
  \begin{minipage}[t][11pt][t]{\linewidth}
    \caption{Results on the LLIE task.}\label{table:low-light}%
  \end{minipage}\par\vspace{0pt}%
  \input{sec/layout_cells/main_table_5.tex}  \end{minipage}\par
\vspace{-4mm}
\end{GRUNPageFloat}

%% file: sec/layout_cells/main_figure_4.tex
\includegraphics[width=1\linewidth]{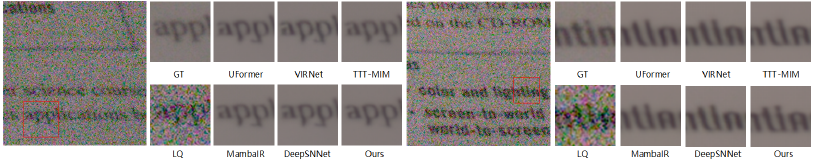}

%% file: sec/layout_cells/main_figure_5.tex
\includegraphics[width=1\linewidth]{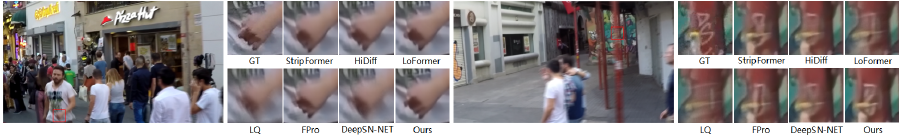}

%% file: sec/layout_cells/main_table_1.tex
{\fontsize{7.2}{8}\selectfont
\setlength{\tabcolsep}{1pt}\renewcommand{\arraystretch}{0.90}%
\begin{tabular*}{\linewidth}{@{\extracolsep{\fill}}l|c|cc|cc}
\toprule
\multirow{2}{*}{Methods} & \multirow{2}{*}{Sources} & \multicolumn{2}{c|}{\textit{SIDD}} & \multicolumn{2}{c}{\textit{DND}} \\ \cline{3-6}
&  & \cellcolor{gray!40}\GRUNCellBand[gray!40]{PSNR~$\uparrow$} & \cellcolor{gray!40}\GRUNCellBand[gray!40]{SSIM~$\uparrow$}       & \cellcolor{gray!40}\GRUNCellBand[gray!40]{PSNR~$\uparrow$} & \cellcolor{gray!40}SSIM~$\uparrow$     \\ \midrule
ADFNet~\cite{shen2023adaptive}                   & AAAI23                   & 39.63       & 0.958      & 39.87      & 0.955      \\
VIRNet~\cite{yue2024deep}                   & TPAMI24                   & 39.64       & 0.958      & 39.83      & 0.954      \\
MambaIR~\cite{guo2024mambair}                  & ECCV24                   &39.89       &0.960       & 40.04      &0.956      \\
TTT-MIM~\cite{mansour2024ttt}                  & ECCV24                   & 39.69       & ---          & 37.04      & ---          \\
DeepSN-Net~\cite{deng2025deepsn}               & TPAMI25                   & 39.79       & 0.958      & 39.92      & 0.956      \\
DnLUT~\cite{yang2025dnlut}                    & CVPR25                   & ---           & 0.875      & 36.67      & 0.922      \\
NLformer~\cite{he2026bayesian} & ECCV26 & 40.11 & 0.960 & 40.08 & 0.956 \\
\rowcolor{c2!20}\GRUNRowBand GRUN               & Ours                     &\textbf{40.19}       &\textbf{0.966}      &    \textbf{40.17}        &   \textbf{0.964}       \\ \bottomrule 
\end{tabular*}%
}

%% file: sec/layout_cells/main_table_2.tex
{\fontsize{7.2}{8}\selectfont
\setlength{\tabcolsep}{1pt}\renewcommand{\arraystretch}{0.90}%
\begin{tabular*}{\linewidth}{@{\extracolsep{\fill}}l|c|cc}
\toprule
 && \multicolumn{2}{c}{\textit{UIEB}}\\ \cline{3-4}
\multirow{-2}{*}{Methods} &\multirow{-2}{*}{Sources} & \cellcolor{gray!40}\GRUNCellBand[gray!40]{PSNR~$\uparrow$} & \cellcolor{gray!40}SSIM~$\uparrow$ \\ \midrule
PUGAN~\cite{cong2023pugan}& TIP23& { {23.05}} & 0.897   \\
ADP~\cite{zhou2023underwater}& IJCV23   & 22.90                                 & 0.892                          \\
NU2Net~\cite{guo2023underwater}   & AAAI23 & 22.38                                 & 0.903      \\
MambaIR~\cite{guo2024mambair} &ECCV24 &22.60 &{ {0.916}}  \\
Reti-Diff~\cite{he2023reti}              & ICLR25        & 24.12 & 0.910 \\ 
UnfoldIR~\cite{he2025unfoldir} & CVPR26 & 24.16 & {0.930}  \\ 
IQPIR~\cite{xiao2026iqpir} & CVPR26 & 24.35 & 0.918 \\
\rowcolor{c2!20}\GRUNRowBand GRUN & Ours &\textbf{24.82} &\textbf{0.943}
\\\bottomrule
\end{tabular*}%
}

%% file: sec/layout_cells/main_table_3.tex
{\fontsize{7.2}{8}\selectfont
\setlength{\tabcolsep}{1pt}\renewcommand{\arraystretch}{0.90}%
\begin{tabular*}{\linewidth}{@{\extracolsep{\fill}}l|c|cc}
\toprule
&  & \multicolumn{2}{c}{\textit{BAID}}                                                                                                                             \\ \cline{3-4}
\multirow{-2}{*}{Methods} & \multirow{-2}{*}{Sources} & \cellcolor{gray!40}\GRUNCellBand[gray!40]{PSNR~$\uparrow$} & \cellcolor{gray!40}SSIM~$\uparrow$ \\ \midrule
Diff-Retinex~\cite{yi2023diff}              & ICCV23                      &22.07 & 0.861     \\
DiffIR~\cite{xia2023diffir}                    & ICCV23                      & 21.10                                 & 0.835   \\
AST~\cite{Zhou_2024_CVPR}  & CVPR24 &22.61 & 0.851  \\
RAVE~\cite{gaintseva2024rave} & ECCV24 & 21.26 & 0.872  \\
Reti-Diff~\cite{he2023reti}  & ICLR25                    & 23.19 & 0.876 \\
UnfoldIR~\cite{he2025unfoldir} & CVPR26 & {24.83} & {0.890}  \\ 
IQPIR~\cite{xiao2026iqpir} & CVPR26 & 24.06 & 0.885 \\
\rowcolor{c2!20}\GRUNRowBand GRUN & Ours &\textbf{24.98} &\textbf{0.907} \\ 
\bottomrule
\end{tabular*}%
}

%% file: sec/layout_cells/main_table_4.tex
{\fontsize{7.2}{8}\selectfont
\setlength{\tabcolsep}{1pt}\renewcommand{\arraystretch}{0.90}%
\begin{tabular*}{\linewidth}{@{\extracolsep{\fill}}l|c|cc|cc}
\toprule
\multirow{2}{*}{Methods} & \multirow{2}{*}{Sources} & \multicolumn{2}{c|}{\textit{GoPro}} & \multicolumn{2}{c}{\textit{HIDE}} \\ \cline{3-6}
& & \cellcolor{gray!40}\GRUNCellBand[gray!40]{PSNR~$\uparrow$}  & \cellcolor{gray!40}\GRUNCellBand[gray!40]{SSIM~$\uparrow$}       & \cellcolor{gray!40}\GRUNCellBand[gray!40]{PSNR~$\uparrow$}  & \cellcolor{gray!40}SSIM~$\uparrow$      \\ \midrule
HI-Diff~\cite{chen2023hierarchical}                & NeurIPS23                   & 33.33       & 0.964      & 31.46      & 0.945      \\
MISC Filter~\cite{liu2024motion}              & CVPR24                   & 34.10       &0.969      & 31.66       & 0.946      \\
FPro~\cite{zhou2024seeing}                     & ECCV24                   & 33.05         & 0.961       & 30.63       & 0.936      \\
LoFormer~\cite{mao2024loformer}                 & ACMM24                   & 34.09       & 0.969        & 31.86        & \textbf{0.949}      \\
DeepSN-Net~\cite{deng2025deepsn}               & TPAMI25                   & 32.83       & 0.960        & 31.14       & 0.941      \\
MDT~\cite{chen2025polarization}                      & CVPR25                   & 34.26       &0.969       & 31.84       & 0.948      \\
LAformer~\cite{ai2026fast} & ECCV26 & 33.40 & 0.965 & 31.49 & 0.945 \\
\rowcolor{c2!20}\GRUNRowBand GRUN                & Ours                     &\textbf{34.30}       &\textbf{0.970}       & \textbf{31.88}       &\textbf{0.949}     \\ \bottomrule
\end{tabular*}%
}

%% file: sec/layout_cells/main_table_5.tex
{\fontsize{7.2}{8}\selectfont
\setlength{\tabcolsep}{1pt}\renewcommand{\arraystretch}{0.90}%
\begin{tabular*}{\linewidth}{@{}l|c|@{\extracolsep{\fill}\hspace{\tabcolsep}}cc|cc|cc}
\toprule
&  & \multicolumn{2}{c|}{\textit{LOL-v1}}& \multicolumn{2}{c|}{\textit{LOL-v2-real}}& \multicolumn{2}{c}{\textit{LOL-v2-synthetic}}\\ \cline{3-8}
\multirow{-2}{*}{Methods} & \multirow{-2}{*}{Sources} 
& \cellcolor{gray!40}\GRUNCellBand[gray!40]{PSNR~$\uparrow$}  & \cellcolor{gray!40}\GRUNCellBand[gray!40]{SSIM~$\uparrow$}  
& \cellcolor{gray!40}\GRUNCellBand[gray!40]{PSNR~$\uparrow$}  & \cellcolor{gray!40}\GRUNCellBand[gray!40]{SSIM~$\uparrow$}  
& \cellcolor{gray!40}\GRUNCellBand[gray!40]{PSNR~$\uparrow$}  & \cellcolor{gray!40}SSIM~$\uparrow$  \\ \midrule
CUE~\cite{zheng2023empowering} & ICCV23 & 21.86 & 0.841 & 21.19 & 0.829 & 24.41 & 0.917 \\
GSAD~\cite{jinhui2023global} & NeurIPS23 & 23.23 & 0.852 & 20.19 & 0.847 &24.22 & 0.927 \\
MambaIR~\cite{guo2024mambair} & ECCV24 & 22.23 & 0.863 & 21.15 & 0.857  & 25.75 & 0.937 \\
Reti-Diff~\cite{he2023reti} & ICLR25 & {25.35} & 0.866 & 22.97 & 0.858 & 27.53 &0.951  \\
CIDNet~\cite{yan2024you} & CVPR25 & 23.50 & 0.900  & {\textbf{24.11}} & 0.871 & 25.71 & 0.942 \\
DCASR~\cite{zhang2026dcasr} & ECCV26 & 24.04 & 0.850 & {24.05} & 0.874 & 25.61 & 0.940 \\
UnfoldIR~\cite{he2025unfoldir} & CVPR26 & 24.41 & {0.911}   & 22.99  & {0.887}  & {27.55} & { 0.959} \\
\rowcolor{c2!20}\GRUNRowBand GRUN & Ours & {\textbf{25.49}} &\textbf{0.914} & 23.83 &\textbf{0.891} &\textbf{27.89} & \textbf{0.961}
\\ \bottomrule
\end{tabular*}%
}

%% file: sec/grun_page7.tex
\begin{GRUNPageFloat}{7}

\noindent
  \begin{minipage}[t]{\linewidth}\vspace{0pt}\centering
  \captionsetup*{type=figure,position=bottom,skip=0.5pt,hypcap=false}%
  \setcounter{figure}{\numexpr6-1\relax}%
  \input{sec/layout_cells/main_figure_6.tex}\par
  \caption{Visual comparisons on BIE,  Dehazing, Deraining, LLIE, and UIE.}\label{Fig.LOL_backlit_underwater}%
  \end{minipage}\par
\vspace{2pt}

\noindent
  \begin{minipage}[t]{\linewidth}\vspace{0pt}\centering
  \captionsetup*{type=table,position=top,skip=0pt}%
  \setcounter{table}{\numexpr6-1\relax}%
  \begin{minipage}[t][21pt][t]{\linewidth}
    \caption{Effect of coupling formulations and cross-factor cues. P+Q/P+G: product coupling with quotient/generic cues; T+M: task-specific coupling with matched cues. Quality: PSNR/SSIM. MI/Grad. Corr.: w/o DCR $\rightarrow$ full GRUN.}\label{tab:coupling}%
  \end{minipage}\par\vspace{0pt}%
  \input{sec/layout_cells/main_table_6.tex}  \end{minipage}\par
\vspace{2pt}

\noindent
  \begin{minipage}[t]{\linewidth}\vspace{0pt}\centering
  \captionsetup*{type=table,position=top,skip=0pt}%
  \setcounter{table}{\numexpr7-1\relax}%
  \begin{minipage}[t][11pt][t]{\linewidth}
    \caption{Results on deraining (Resolution: $256\times 256$).}\label{table:deraining}%
  \end{minipage}\par\vspace{0pt}%
  \input{sec/layout_cells/main_table_7.tex}  \end{minipage}\par
\vspace{2pt}

\noindent
  \begin{minipage}[t]{.49\linewidth}\vspace{0pt}\centering
  \captionsetup*{type=table,position=top,skip=0pt}%
  \setcounter{table}{\numexpr8-1\relax}%
  \begin{minipage}[t][11pt][t]{\linewidth}
    \caption{Results on the real-world IDIR task.}\label{table:realIDIR}%
  \end{minipage}\par\vspace{0pt}%
  \input{sec/layout_cells/main_table_8.tex}  \end{minipage}
\hfill
  \begin{minipage}[t]{.49\linewidth}\vspace{0pt}\centering
  \captionsetup*{type=table,position=top,skip=0pt}%
  \setcounter{table}{\numexpr9-1\relax}%
  \begin{minipage}[t][11pt][t]{\linewidth}
    \caption{Results on the dehazing task.}\label{table:dehazing}%
  \end{minipage}\par\vspace{0pt}%
  \input{sec/layout_cells/main_table_9.tex}  \end{minipage}\par
\vspace{-4mm}
\end{GRUNPageFloat}

%% file: sec/layout_cells/main_figure_6.tex
\includegraphics[width=\linewidth]{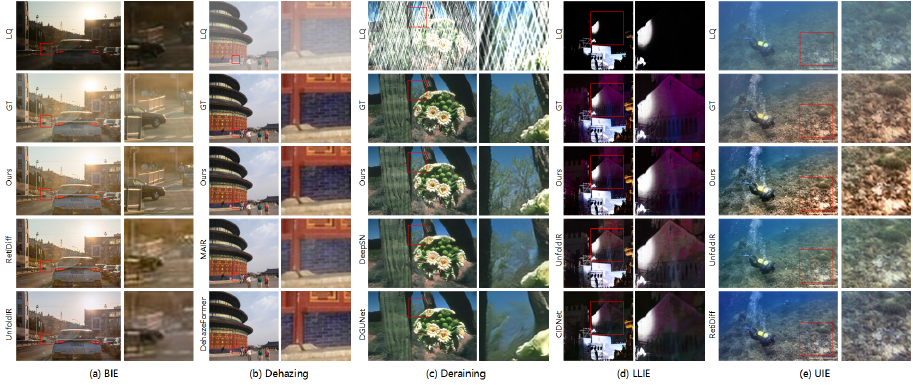}

%% file: sec/layout_cells/main_table_6.tex
{\fontsize{8}{9.2}\selectfont
\setlength{\tabcolsep}{1.5pt}\renewcommand{\arraystretch}{0.9}%
\begin{tabular*}{\linewidth}{@{\extracolsep{\fill}}lccccc@{}}
\toprule
\textbf{Dataset} & MI $\downarrow$ & Grad. Corr. $\downarrow$ & \textbf{T+M} & \textbf{P+G} & \cellcolor{c2!20}\textbf{P+Q} \\
\midrule
SIDD & 0.19 $\rightarrow$ \textbf{0.08} & 0.27 $\rightarrow$ \textbf{0.15} & 39.98/0.962 & 39.98/0.963 & \cellcolor{c2!20}\textbf{40.19/0.966} \\
Rain100L & 0.22 $\rightarrow$ \textbf{0.13} & 0.31 $\rightarrow$ \textbf{0.17} & 39.18/0.980 & 39.34/0.981 & \cellcolor{c2!20}\textbf{39.64/0.984} \\
GoPro & 0.26 $\rightarrow$ \textbf{0.14} & 0.34 $\rightarrow$ \textbf{0.21} & 34.12/0.968 & 34.11/0.966 & \cellcolor{c2!20}\textbf{34.30/0.970} \\
SOTS-Outdoor & 0.29 $\rightarrow$ \textbf{0.16} & 0.37 $\rightarrow$ \textbf{0.23} & 38.05/0.991 & 37.96/0.990 & \cellcolor{c2!20}\textbf{38.13/0.992} \\
UIEB & 0.18 $\rightarrow$ \textbf{0.11} & 0.26 $\rightarrow$ \textbf{0.18} & 24.66/0.939 & 24.63/0.939 & \cellcolor{c2!20}\textbf{24.82/0.943} \\
\bottomrule
\end{tabular*}%
}

%% file: sec/layout_cells/main_table_7.tex
\renewcommand{\arraystretch}{0.86}%
\resizebox{\linewidth}{!}{%
\setlength{\tabcolsep}{1.6mm}%
\begin{tabular}{l|c|cc|cc|cc|cc|cc|cc}
\toprule
&  &Para.  & FLOPs &\multicolumn{2}{c|}{\textit{Rain100L}} & \multicolumn{2}{c|}{\textit{Rain100H}}                                                     & \multicolumn{2}{c|}{\textit{Test100}}                                                   & \multicolumn{2}{c|}{\textit{Test1200}}                                                  & \multicolumn{2}{c}{\textit{Test2800}}                                                  \\ \cline{5-14}
\multirow{-2}{*}{Methods} & \multirow{-2}{*}{Sources} & (M) & (G) & \cellcolor{gray!40}PSNR~$\uparrow$  & \cellcolor{gray!40}SSIM~$\uparrow$ & \cellcolor{gray!40}PSNR~$\uparrow$  & \cellcolor{gray!40}SSIM~$\uparrow$& \cellcolor{gray!40}PSNR~$\uparrow$  & \cellcolor{gray!40}SSIM~$\uparrow$& \cellcolor{gray!40}PSNR~$\uparrow$  & \cellcolor{gray!40}SSIM~$\uparrow$& \cellcolor{gray!40}PSNR~$\uparrow$  & \cellcolor{gray!40}SSIM~$\uparrow$\\ \midrule
DGUNet~\cite{mou2022deep} & CVPR22 & 69.57 & 2335.46 & 37.42 & 0.969 & 30.66 & 0.891 & 30.32 & 0.899 &33.23 & 0.920 & 33.68 & 0.938  \\
IR-SDE~\cite{luo2023image}                    & ICML23      &  135.30  &    469.23            & 38.30                                 & {0.980} & 31.65                                 & 0.904                                 & ---                                     & ---                                     & ---                                     & ---                                     & 30.42                                 & 0.891                                 \\
MFDNet~\cite{wang2023multi}                    &  TIP23      &  ---  &    ---            & 37.61                                 & 0.973 & 30.48                                 & 0.899                                 & 30.78                                     & 0.914                                     & 33.01                                     & 0.925                                     & 33.55                                 & 0.939                                                                                              \\
DAWN~\cite{jiang2023dawn}                    &  ACMM23      &  ---  &    ---            & 35.97                              & 0.963  &  29.89                                 & 0.889                                 & 29.86                                    &  0.902                                    & 32.76                                    &  0.919                                     & ---                                 & ---                                                                                              \\
MambaIR~\cite{guo2024mambair}                   & ECCV24      & 31.51  &  363.84                        & 38.78                                 & 0.977                                 & 30.62                                 & 0.893                                 & ---                                     & ---                                     & 32.56                                 & 0.923                                 & 33.58                                 & 0.927                                 \\
VMambaIR~\cite{shi2025vmambair}                  & TCSVT25      & 25.17   &  537.26                       & {39.09} & 0.979                                 & 31.66                                 & {0.909} & ---                                     & ---                                     & 33.33                                 & 0.926                                 & 34.01                                 & {0.944} \\
DeepSN-Net~\cite{deng2025deepsn}                & TPAMI25     &  22.72  & 364.13                        & 38.59                                 & 0.975                                 & {31.81} & 0.904                                 &31.60                                 & {0.920} & {33.45} & {0.931} &34.01                                 & 0.942                                 \\
\rowcolor{c2!20} GRUN                 & Ours         &  21.54  &  83.87                       & {\textbf{39.64}} & {\textbf{0.984}} & {\textbf{32.33}} & {\textbf{0.915}} & {\textbf{32.74}} & {\textbf{0.926}} & {\textbf{34.25}} & {\textbf{0.939}} & {\textbf{34.35}} & {\textbf{0.951}} \\ \bottomrule
\end{tabular}%
}

%% file: sec/layout_cells/main_table_8.tex
\renewcommand{\arraystretch}{0.86}%
\resizebox{\linewidth}{!}{%
\setlength{\tabcolsep}{1mm}%
\begin{tabular}{l|cc|cc|cc}
\toprule
\multirow{2}{*}{Methods}& \multicolumn{2}{c|}{\textit{DICM}} & \multicolumn{2}{c|}{\textit{VV}} & \multicolumn{2}{c}{\textit{MEF}}  \\
&\cellcolor{gray!40}PI~$\downarrow$       & \cellcolor{gray!40}NIQE~$\downarrow$              & \cellcolor{gray!40}PI~$\downarrow$       & \cellcolor{gray!40}NIQE~$\downarrow$              & \cellcolor{gray!40}PI~$\downarrow$      & \cellcolor{gray!40}NIQE~$\downarrow$                 \\ \midrule
PairLIE~\cite{fu2023learning} & {3.685}  & 4.034&   3.334 & 3.574 &  4.133  & 4.065  \\
Reti-Diff~\cite{he2023reti} &2.351     &3.255    & 2.651     & 2.540    &  {3.308} & {3.792}   \\ 
CIDNet~\cite{yan2024you}  &
3.045 & 3.796 & 2.826  & 3.218 &
2.683 & 3.568   \\
UnfoldIR~\cite{he2025unfoldir} & {2.952} & 3.381 & 2.553 & 2.306 & 2.722 & 3.387 \\
\rowcolor{c2!20} GRUN & \textbf{2.265} & \textbf{3.147} & \textbf{2.521} & \textbf{2.287} & \textbf{2.524} & \textbf{3.145} \\
\bottomrule
\end{tabular}%
}

%% file: sec/layout_cells/main_table_9.tex
\renewcommand{\arraystretch}{0.86}%
\resizebox{\linewidth}{!}{%
\setlength{\tabcolsep}{1.5mm}%
\begin{tabular}{l|cc|cc}
\toprule
\multirow{2}{*}{Methods} 
& \multicolumn{2}{c|}{\textit{SOTS-Indoor}} 
& \multicolumn{2}{c}{\textit{SOTS-Outdoor}} \\ \cline{2-5}

& \cellcolor{gray!40}PSNR$\uparrow$ 
& \cellcolor{gray!40}SSIM$\uparrow$ 
& \cellcolor{gray!40}PSNR$\uparrow$ 
& \cellcolor{gray!40}SSIM$\uparrow$ \\

\midrule

MB-TaylorFormer~\cite{qiu2023mb} 
& 40.71 & 0.992 
& 37.42 & 0.989 \\

C2PNet~\cite{zheng2023curricular} 
& 42.56 & 0.996 
& 36.68 & 0.990 \\

DCMPNet~\cite{zhang2024depth} 
& 42.18 & \textbf{0.997} 
& 36.56 & \textbf{0.993} \\

MaIR~\cite{li2025mair} 
& 39.45 & \textbf{0.997} 
& 36.96 & 0.991 \\

\rowcolor{c2!20}
GRUN
& \textbf{42.60} 
& \textbf{0.997} 
& \textbf{38.13} 
& 0.992 \\

\bottomrule
\end{tabular}%
}

%% file: sec/grun_page8.tex
\begin{GRUNPageFloat}{8}

\noindent
  \begin{minipage}[t]{.49\linewidth}\vspace{0pt}\centering
  \captionsetup*{type=figure,position=bottom,skip=0.5pt,hypcap=false}%
  \setcounter{figure}{\numexpr7-1\relax}%
  \input{sec/layout_cells/main_figure_7.tex}\par
  \caption{Visualization of the decomposition.}\label{fig:decomposition}%
  \end{minipage}
\hfill
  \begin{minipage}[t]{.49\linewidth}\vspace{0pt}\centering
  \captionsetup*{type=figure,position=bottom,skip=0.5pt,hypcap=false}%
  \setcounter{figure}{\numexpr8-1\relax}%
  \input{sec/layout_cells/main_figure_8.tex}\par
  \caption{Cross-stage visualization.}\label{fig:cross_stage}%
  \end{minipage}\par
\vspace{2pt}

\noindent
  \begin{minipage}[t]{\linewidth}\vspace{0pt}\centering
  \captionsetup*{type=table,position=top,skip=0pt}%
  \setcounter{table}{\numexpr10-1\relax}%
  \begin{minipage}[t][11pt][t]{\linewidth}
    \caption{Ablation study of our GRUN.}\label{table:Ablation}%
  \end{minipage}\par\vspace{0pt}%
  \input{sec/layout_cells/main_table_10.tex}  \end{minipage}\par
\vspace{2pt}
\input{sec/grun_table11_combined}
\end{GRUNPageFloat}

%% file: sec/layout_cells/main_figure_7.tex
\includegraphics[width=\linewidth,height=4.2cm,keepaspectratio]{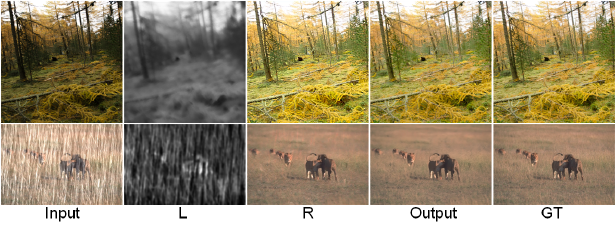}

%% file: sec/layout_cells/main_figure_8.tex
\includegraphics[width=\linewidth,height=4.2cm,keepaspectratio]{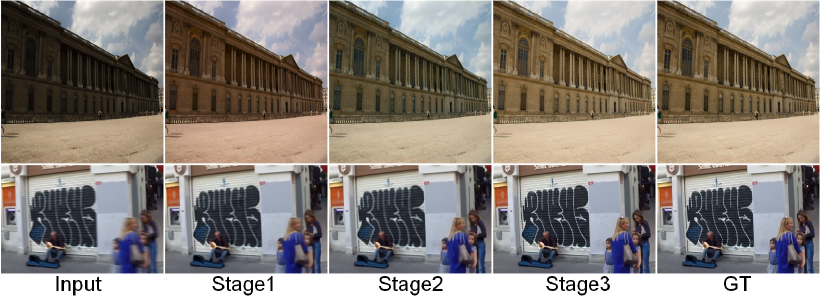}

%% file: sec/layout_cells/main_table_10.tex
\renewcommand{\arraystretch}{0.86}%
\resizebox{\linewidth}{!}{%
\setlength{\tabcolsep}{1mm}%
\begin{tabular}{l|c|ccccccc|cccc|c|c}
\toprule
\multirow{2}{*}{Datasets} & \multirow{2}{*}{Metrics} 
& \multicolumn{7}{c|}{Generalized Retinex Decomposition} 
& \multicolumn{4}{c|}{Degradation Estimation}
& \multicolumn{1}{c|}{DCR}
& \cellcolor{c2!20} GRUN \\
\cline{3-15}
& 
& w/o Retinex 
& w/o $z^{(k)}$ 
& w/o $R^{(k)}$ 
& w/o $L^{(k)}$ 
& $\mathcal{L}_1(\cdot)$ 
& $\mathcal{L}_2(\cdot)$ 
& w/o $\mathcal{F_{\theta}}(\cdot)$ 
& w/o DE 
& w/o $\mathbf{c}$  
& $\mathcal{D}_1(\cdot)$ 
& $\mathcal{D}_2(\cdot)$ 
& w/o DCR 
& \cellcolor{c2!20} Ours \\
\midrule

\multirow{2}{*}{\textit{L-v2-r}} 
& PSNR 
& 22.72 & 23.48 & 22.93 & 22.83 & 23.24 & 23.38 & 23.47
& 23.10 & 23.34 & 23.22 & 23.28
& 23.37
& \cellcolor{c2!20}\textbf{23.83} \\
& SSIM
& 0.854 & 0.880 & 0.861 & 0.858 & 0.872 & 0.873 & 0.882
& 0.869 & 0.878 & 0.874 & 0.876
& 0.876
& \cellcolor{c2!20}\textbf{0.891} \\
\midrule

\multirow{2}{*}{\textit{L-v2-s}} 
& PSNR
& 26.27 & 27.68 & 26.74 & 26.55 & 27.06 & 27.03 & 27.41
& 27.01 & 27.26 & 27.14 & 27.19
& 27.39
& \cellcolor{c2!20}\textbf{27.89} \\
& SSIM
& 0.922 & 0.950 & 0.930 & 0.925 & 0.943 & 0.944 & 0.952
& 0.937 & 0.944 & 0.941 & 0.943
& 0.947
& \cellcolor{c2!20}\textbf{0.961} \\
\bottomrule
\end{tabular}%
}

%% file: sec/grun_table11_combined.tex
\begin{minipage}[t]{\linewidth}\vspace{0pt}\centering
\captionsetup*{type=table,position=top,skip=0pt}
\setcounter{table}{10}
\caption{Cross-architecture applicability of GRUN components. ``++'' marks the adapted models.}\label{table:generalization}
\input{sec/layout_cells/main_table_11_combined}
\end{minipage}

\par\vspace{2pt}
\noindent
\begin{minipage}[t]{.72\linewidth}\vspace{0pt}\captionsetup*{type=table,position=top,skip=0pt}\setcounter{table}{11}\vspace{0pt}\centering
\begin{minipage}[t][10.5pt][t]{\linewidth}\caption{Low-light image detection on ExDark}\label{table:Detection}\end{minipage}\par\vspace{0pt}
{\fontsize{7}{8}\selectfont
\setlength{\tabcolsep}{1pt}\renewcommand{\arraystretch}{0.86}%
\begin{tabular*}{\linewidth}{@{\extracolsep{\fill}}l|cccccccccccc|c}
\toprule
Methods (AP)
& Bicycle
& Boat
& Bottle
& Bus
& Car
& Cat
& Chair
& Cup
& Dog
& Motor
& People
& Table
& Mean \\
\midrule
Baseline
& 74.7 & 64.9 & 70.7 & 84.2 & 79.7 & 47.3 & 58.6 & 67.1 & 64.1 & 66.2 & 73.9 & 45.7 & 66.4 \\
SCI~\cite{ma2022toward}
& 73.4 & 68.0 & 69.5 & 86.2 & 74.5 & 63.1 & 59.5 & 61.0 & 67.3 & 63.9 & 73.2 & 47.3 & 67.2 \\
SNR-Net~\cite{xu2022snr}
& 78.3 & 74.2 & 74.5 & 89.6 & 82.7 & 66.8 & 66.3 & 62.5 & 74.7 & 63.1 & 73.3 & 57.2 & 71.9 \\
Reti-Diff~\cite{he2023reti}
& 82.0 & 77.9 & 76.4 & {\textbf{92.2}} & {83.3}
& 69.6 & 67.4 & 74.4 & 75.5 & 74.3 & 78.3 & 57.9 & 75.8\\
UnfoldIR~\cite{he2025unfoldir}
& {87.5} & {81.0}
& {78.2} & 86.4 & 74.6
& {76.4} & {80.1}
& {\textbf{80.8}} & {\textbf{83.4}}
& {84.1} & {79.1}
& {65.0} & {79.7} \\
\rowcolor{c2!20}\GRUNRowBand GRUN
& {\textbf{88.5}}
& {\textbf{81.4}}
& {\textbf{79.1}}
& {91.4}
& {\textbf{83.4}}
& {\textbf{76.8}}
& {\textbf{80.5}}
& {77.8}
& {81.6}
& {\textbf{85.4}}
& {\textbf{79.5}}
& {\textbf{65.9}}
& {\textbf{80.9}} \\
\bottomrule
\end{tabular*}%
}

\end{minipage}
\hfill
\begin{minipage}[t]{.265\linewidth}\vspace{0pt}\setcounter{table}{12}\centering
\captionsetup*{type=table,position=top,skip=0pt}
\begin{minipage}[t][10.5pt][t]{\linewidth}\caption{Effect of explicit coupling.}\label{tab:retinex_pathway}\end{minipage}\par\vspace{0pt}
{\fontsize{7}{8}\selectfont
\setlength{\tabcolsep}{1pt}\renewcommand{\arraystretch}{1.1325}%
\begin{tabularx}{\linewidth}{l|>{\centering\arraybackslash}X|>{\centering\arraybackslash}X}
\toprule
\rule[-6pt]{0pt}{18pt}\textbf{Configuration} & \makecell{\textbf{GoPro}\\PSNR/SSIM} & \makecell{\textbf{SIDD}\\PSNR/SSIM} \\
\midrule
\makecell[l]{Generic\\Two-Branch} & 32.86/0.958 & 38.97/0.949 \\
\midrule
\rowcolor{c2!20}\textbf{Full GRUN} & \textbf{34.30/0.970} & \textbf{40.19/0.966} \\
\bottomrule
\end{tabularx}%
}
\end{minipage}\par
\setcounter{table}{13}
\vspace{-6mm}

%% file: sec/layout_cells/main_table_11_combined.tex
\noindent
\begin{minipage}[t]{.32\linewidth}\vspace{0pt}\setcounter{table}{11}\begin{subtable}[t]{\linewidth}\setcounter{subtable}{0}\vspace{0pt}\centering
\begin{minipage}[t][10.5pt][t]{\linewidth}\captionsetup{margin=0pt,justification=centering,singlelinecheck=false}\caption{Image deblurring}\end{minipage}\par\vspace{0pt}
{\fontsize{7}{8}\selectfont
\setlength{\tabcolsep}{1pt}\renewcommand{\arraystretch}{0.94}%
\begin{tabularx}{\linewidth}{@{}l|w{c}{32.36pt}|>{\centering\arraybackslash}Xw{c}{34.91pt}}
\toprule
\rule[-6pt]{0pt}{18pt}Datasets & Metrics
& DGUNet~\cite{mou2022deep}
& \cellcolor{c2!20} DGUNet++ \\
\midrule
\multirow{2}{*}{\textit{GoPro}}
& PSNR$\uparrow$ & 32.71 & \cellcolor{c2!20}\textbf{33.59} \\
& SSIM$\uparrow$ & 0.960 & \cellcolor{c2!20}\textbf{0.964} \\
\midrule
\multirow{2}{*}{\textit{HIDE}}
& PSNR$\uparrow$ & 30.96 & \cellcolor{c2!20}\textbf{31.57} \\
& SSIM$\uparrow$ & 0.940 & \cellcolor{c2!20}\textbf{0.947} \\
\bottomrule
\end{tabularx}%
}

\end{subtable}
\end{minipage}
\hfill
\begin{minipage}[t]{.32\linewidth}\vspace{0pt}\setcounter{table}{11}\begin{subtable}[t]{\linewidth}\setcounter{subtable}{1}\vspace{0pt}\centering
\begin{minipage}[t][10.5pt][t]{\linewidth}\caption{Image deraining}\end{minipage}\par\vspace{0pt}
{\fontsize{7}{8}\selectfont
\setlength{\tabcolsep}{1pt}\renewcommand{\arraystretch}{0.94}%
\begin{tabular*}{\linewidth}{@{\extracolsep{\fill}}l|c|cc}
\toprule
\rule[-6pt]{0pt}{18pt}Datasets & Metrics
& DeepSN-Net~\cite{deng2025deepsn}
& \cellcolor{c2!20} DeepSN-Net++ \\
\midrule
\multirow{2}{*}{\textit{Rain100L}}
& PSNR$\uparrow$ & 38.59 & \cellcolor{c2!20}\textbf{38.92} \\
& SSIM$\uparrow$ & 0.975 & \cellcolor{c2!20}\textbf{0.980} \\
\midrule
\multirow{2}{*}{\textit{Rain100H}}
& PSNR$\uparrow$ & 31.81 & \cellcolor{c2!20}\textbf{32.03} \\
& SSIM$\uparrow$ & 0.904 & \cellcolor{c2!20}\textbf{0.911} \\
\bottomrule
\end{tabular*}%
}

\end{subtable}
\end{minipage}
\hfill
\begin{minipage}[t]{.32\linewidth}\vspace{0pt}\setcounter{table}{11}\begin{subtable}[t]{\linewidth}\setcounter{subtable}{2}\vspace{0pt}\centering
\begin{minipage}[t][10.5pt][t]{\linewidth}\caption{Image dehazing}\end{minipage}\par\vspace{0pt}
{\fontsize{7}{8}\selectfont
\setlength{\tabcolsep}{1pt}\renewcommand{\arraystretch}{1.026}%
\begin{tabular*}{\linewidth}{@{\extracolsep{\fill}}l|c|cc}
\toprule
\rule[-6pt]{0pt}{18pt}Datasets & Metrics
& CORUN~\cite{fang2024real}
& \cellcolor{c2!20} CORUN++ \\
\midrule
\multirow{4}{*}{\textit{RTTS}}
& FADE$\downarrow$ & 0.824 & \cellcolor{c2!20}\textbf{0.801} \\
& BRISQUE$\downarrow$ & 11.96 & \cellcolor{c2!20}\textbf{10.59} \\
& MUSIQ-K$\uparrow$ & 63.82 & \cellcolor{c2!20}\textbf{68.97} \\
& CLIP-IQA$\uparrow$ & 0.683 & \cellcolor{c2!20}\textbf{0.746} \\
\bottomrule
\end{tabular*}%
}

\end{subtable}
\end{minipage}\par
\vspace{2pt}

\noindent
\begin{minipage}[t]{.32\linewidth}\vspace{0pt}\setcounter{table}{11}\begin{subtable}[t]{\linewidth}\setcounter{subtable}{3}\vspace{0pt}\centering
\begin{minipage}[t][10.5pt][t]{\linewidth}\caption{Underwater image enhancement}\end{minipage}\par\vspace{0pt}
{\fontsize{7}{8}\selectfont
\setlength{\tabcolsep}{1pt}\renewcommand{\arraystretch}{0.94}%
\begin{tabularx}{\linewidth}{c|c|>{\centering\arraybackslash}X>{\columncolor{c2!20}\centering\arraybackslash}X}
\toprule
\rule[-3pt]{0pt}{11pt}Datasets & Metrics & UnfoldIR~\cite{he2025unfoldir} & UnfoldIR++ \\
\midrule
\multirow{2}{*}{\textit{UIEB}}
& PSNR~$\uparrow$ & 24.16 & \textbf{24.48} \\
& SSIM~$\uparrow$ & 0.930 & \textbf{0.937} \\
\bottomrule
\end{tabularx}%
}
\end{subtable}
\end{minipage}
\hfill
\begin{minipage}[t]{.32\linewidth}\vspace{0pt}\setcounter{table}{11}\begin{subtable}[t]{\linewidth}\setcounter{subtable}{4}\vspace{0pt}\centering
\begin{minipage}[t][10.5pt][t]{\linewidth}\caption{Backlit image enhancement}\end{minipage}\par\vspace{0pt}
{\fontsize{7}{8}\selectfont
\setlength{\tabcolsep}{1pt}\renewcommand{\arraystretch}{0.94}%
\begin{tabularx}{\linewidth}{c|c|>{\centering\arraybackslash}X>{\columncolor{c2!20}\centering\arraybackslash}X}
\toprule
\rule[-3pt]{0pt}{11pt}Datasets & Metrics & Reti-Diff~\cite{he2023reti} & Reti-Diff++ \\
\midrule
\multirow{2}{*}{\textit{BAID}}
& PSNR~$\uparrow$ & 23.19 & \textbf{23.62} \\
& SSIM~$\uparrow$ & 0.876 & \textbf{0.888} \\
\bottomrule
\end{tabularx}%
}
\end{subtable}
\end{minipage}
\hfill
\begin{minipage}[t]{.32\linewidth}\vspace{0pt}\setcounter{table}{11}\begin{subtable}[t]{\linewidth}\setcounter{subtable}{5}\vspace{0pt}\centering
\begin{minipage}[t][10.5pt][t]{\linewidth}\caption{Low-light image enhancement}\label{table:cm_ablation}\end{minipage}\par\vspace{0pt}
{\fontsize{7}{8}\selectfont
\setlength{\tabcolsep}{1pt}\renewcommand{\arraystretch}{0.94}%
\begin{tabularx}{\linewidth}{c|c|>{\centering\arraybackslash}X>{\columncolor{c2!20}\centering\arraybackslash}X}
\toprule
\rule[-3pt]{0pt}{11pt}Datasets & Metrics & URetinex-Net~\cite{wu2022uretinex} & URetinex-Net++ \\
\midrule
\multirow{2}{*}{\textit{L-v2-r}}
& PSNR~$\uparrow$ & 20.44 & \textbf{22.14} \\
& SSIM~$\uparrow$ & 0.806 & \textbf{0.845} \\
\bottomrule
\end{tabularx}%
}

\end{subtable}
\end{minipage}\par
\vspace{2pt}
\setcounter{table}{11}